\documentclass[10pt,twocolumn,letterpaper]{article}

\usepackage{cvpr}              

\usepackage{graphicx}
\usepackage{amsmath}
\usepackage{amssymb}
\usepackage{booktabs}
\usepackage{import}
\usepackage{cuted}
\usepackage{lipsum}
\usepackage{microtype}
\usepackage[dvipsnames]{xcolor}
\usepackage{caption}
\usepackage{subcaption}

\usepackage[pagebackref,breaklinks,colorlinks]{hyperref}
\usepackage{tablefootnote}

\usepackage[capitalize]{cleveref}
\crefname{section}{Sec.}{Secs.}
\Crefname{section}{Section}{Sections}
\Crefname{table}{Table}{Tables}
\crefname{table}{Tab.}{Tabs.}

\def\cvprPaperID{4288} 
\def\confName{CVPR}
\def\confYear{2022}
\newcommand{\Hquad}{\hspace{1.5em}} 

\begin{document}

\title{FLAG: Flow-based 3D Avatar Generation from Sparse Observations}

\author{
\fontsize{11pt}{11pt}\selectfont Sadegh Aliakbarian\selectfont
\Hquad 
\fontsize{11pt}{11pt}\selectfont Pashmina Cameron\selectfont
\Hquad 
\fontsize{11pt}{11pt}\selectfont Federica Bogo\selectfont
\Hquad 
\fontsize{11pt}{11pt}\selectfont Andrew Fitzgibbon\selectfont
\Hquad 
\fontsize{11pt}{11pt}\selectfont Thomas J. Cashman\selectfont
\\ [0.5em]
\fontsize{11pt}{11pt}\selectfont Mixed Reality \& AI Lab, Microsoft\selectfont
\\
{\small \tt \url{https://microsoft.github.io/flag}}
}

\maketitle


\begin{abstract}
    To represent people in mixed reality applications for collaboration and communication, we need to generate realistic and faithful avatar poses.
However, the signal streams that can be applied for this task from head-mounted devices (HMDs) are typically limited to head pose and hand pose estimates.
While these signals are valuable, they are an incomplete representation of the human body, making it challenging to generate a faithful full-body avatar.
We address this challenge by developing a flow-based generative model of the 3D human body from sparse observations, wherein we learn not only a conditional distribution of 3D human pose, but also a probabilistic mapping from observations to the latent space from which we can generate a plausible pose along with uncertainty estimates for the joints. 
We show that our approach is not only a strong predictive model, but can also act as an efficient pose prior in different optimization settings where a good initial latent code plays a major role.

\end{abstract}

\section{Introduction}
\label{sec:introduction}

Mixed reality technology provides new ways to interact with people, with applications in remote collaboration, virtual gatherings, gaming and education. 
\emph{People} are at the heart of all these applications, and so generating realistic human representations with high fidelity is key to the user experience. 
Whilst
external sensors and cameras~\cite{saito2020pifuhd} are effective, using only head-mounted devices (HMDs) to generate realistic and faithful human representations remains a challenging problem.
The relevant data available from HMDs such as Microsoft HoloLens and Oculus Quest is limited to the location and orientation of the head and the location and orientation of the hands, obtained either via egocentric hand tracking~\cite{taylor2016efficient,han2020megatrack} or the signal from motion controllers.
This is a very incomplete signal for the pose and motion of the full human body.

\begin{figure}
    \centering
    \includegraphics[width=\columnwidth]{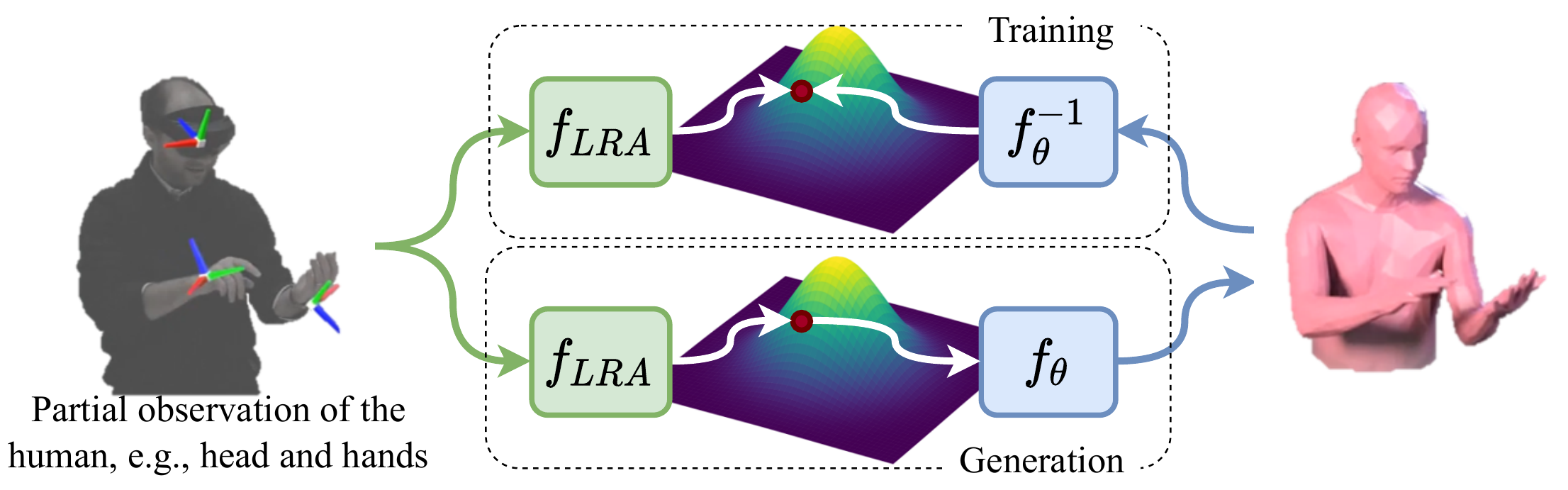}\vspace{-5pt}
    \caption{
    We generate a full body avatar given sparse HMD input (three SE3s for head and hands), by training a flow-based generative model that provides an invertible mapping between the base distribution and 3D human pose distribution.
    At test time, given the HMD signal, we predict a region in the latent space that is used as the input to the flow-based model to generate a pose.}
    \label{fig:teaser}
    \vspace{-10pt}
\end{figure}

Although prior work has proposed human pose priors for generating 3D human body poses from partial and ambiguous observations such as images~\cite{kolotouros2021probabilistic,biggs20203d,zanfir2020weakly}, 2D joints/keypoints~\cite{pavlakos2019expressive,bogo2016keep}, and markers~\cite{zhang2021we,loper2014mosh,zanfir2021thundr,ghorbani2021soma}, such observations are richer sources of data than those available in practice from HMDs.
Despite the importance of this problem, there have been few attempts to generate full body pose from extremely sparse observations, i.e., head and hand position and orientation.
Dittadi et al.~\cite{dittadi2021full} developed a variational autoencoder (VAE) to compress the head and hands inputs to a latent space, allowing a full-body pose to be generated by sampling from that latent space.

We propose a new approach based on conditional normalizing flows for \emph{sparse inputs}. 
Specifically, we learn the conditional distribution of the full body pose given the head and hand data via a flow-based model which enables an invertible mapping between the 3D pose distribution and the base distribution.
Invertibility of our model then allows us to go further by learning a probabilistic mapping from the condition to the \emph{high-likelihood} region in the same base distribution, as illustrated in Fig.~\ref{fig:teaser}.
We name our approach a \textbf{fl}ow-based \textbf{a}vatar \textbf{g}enerative model (\textbf{FLAG}).
The strengths of this design are: first, using a flow-based generative model enables \emph{exact pose likelihood} computation in contrast to the approximate likelihoods seen in VAE-based pose priors~\cite{pavlakos2019expressive,dittadi2021full}.
Second, the invertibility of our generative model allows us to compute the \emph{oracle latent code}. 
During training, the oracle latent code then acts as the ground truth for our mapping function.
This allows us to learn a representative mapping from the observed head and hands to the latent space, making our approach a \emph{strong predictive model}.
Finally, when optimizing either in pose space or latent space, using our model as the pose prior provides a \emph{superior initialization} in the latent space, making optimization very efficient.

\section{Related Work}
\label{sec:relatedwork}
Several recent works generate 3D human body pose given partial observations, such as images~\cite{kolotouros2021probabilistic,biggs20203d,sengupta2021probabilistic,kolotouros2019learning}, 2D keypoints~\cite{bogo2016keep,kanazawa2018end,kolotouros2021probabilistic}, HMDs~\cite{dittadi2021full}, IMUs~\cite{guzov2021human} and additional upper body tracking signals~\cite{yang2021lobstr}, or trajectories of partially visible body joints~\cite{kania2021trajevae}.
These methods usually require richer input than is available from commercial HMDs~\cite{ungureanu2020hololens}, whereas we wish to address the challenge of generating full body poses solely from HMD input. 
Most related work uses a generative model of human pose, either to directly predict the parameters of the body model~\cite{loper2015smpl,osman2020star,pavlakos2019expressive} or as a pose prior in an optimization framework~\cite{bogo2016keep,kolotouros2021probabilistic,zhang2021we}. 
A few authors combine the two, either by training a network that mimics the behavior of an optimizer~\cite{kanazawa2018end,zanfir2021neural} or an optimizer initialised using a neural network~\cite{kolotouros2019learning}. 

SMPLify~\cite{bogo2016keep} proposed a probabilistic 3D human pose prior based on a mixture of Gaussians.
Pavlakos et al.~\cite{pavlakos2019expressive} found SMPLify insufficiently expressive to model the complex human pose distribution and proposed VPoser, which uses variational autoencoders~\cite{kingma2013auto}.
When using unconditional VAE-based pose priors, consistency with the observations has to be enforced via additional terms in the optimization cost function.
In contrast, conditional VAE-based (CVAE)~\cite{sohn2015learning} pose priors use the observations to estimate the pose likelihood.
Liang et al.~\cite{ling2020character} and Rempe et al.~\cite{rempe2021humor} use previously observed poses to condition the pose prior while
Sharma et al.~\cite{sharma2019monocular} use a CVAE to generate 3D human pose from 2D keypoints extracted from images.

Previous studies~\cite{higgins2016beta,razavi2019preventing,bowman2015generating} have established that VAE-based approaches
are challenging to train due to the heuristic nature of tuning the balance between the reconstruction and the KL divergence loss in the VAE's evidence lower bound (ELBO).
If the goal is to learn a rich semantic latent space close to a normal distribution, the weight for the KL term needs to be relatively high (close to 1 as in the standard ELBO), which in turn leads to lower quality pose reconstruction through decoding.
If one requires high-quality pose reconstruction, then the weights for the KL should be relatively small, e.g., $5 \times 10^{-3}$ in VPoser~\cite{pavlakos2019expressive}, leading to a model that does not optimize the true ELBO with an imperfect latent representation.
This push-pull effect on the weights of the two terms makes VAE training difficult, but it becomes even more challenging when strong conditioning signals, such as images, previous poses, or 2D keypoints~\cite{rempe2021humor,sharma2019monocular,ling2020character} are introduced. 
If trained in the standard fashion, the conditioning signal is strong enough that the decoder can generate a pose given only the condition, and thus learns to ignore the latent variable~\cite{aliakbarian2020stochastic,aliakbarian2021contextually}.
To avoid this, CVAE-based pose priors tend to assign a very small weight to the KL term, e.g., $4 \times 10^{-4}$ as in Rempe et al.~\cite{rempe2021humor}, to avoid posterior collapse~\cite{higgins2016beta,razavi2019preventing,bowman2015generating}.

Unlike VAE-based approaches, normalizing flow-based models represent the complex data distribution via a composition of invertible transforms and minimize the \textit{exact} negative log-likelihood of the poses.
Biggs et al.~\cite{biggs20203d} use a flow-based model
as a pose prior in an optimization problem where the goal is find a likely pose that minimizes the re-projection error given 2D keypoints.
Zanfir et al.~\cite{zanfir2020weakly} use a flow-based pose prior to fit a 3D body model on 2D images in a weakly-supervised framework.
Kolotouros et al.~\cite{kolotouros2021probabilistic} extend these models to make them conditional on observed images enabling them to use a single model both as a pose prior and directly as a predictive model, allowing generation of a plausible 3D human pose given an image and a latent code ($z=\mathbf{0}$).
These advances build confidence in highly expressive conditional flow-based models with rich conditional inputs such as images or keypoints.

\noindent\textbf{Our approach}
In this paper, we push this line of investigation even further and propose FLAG, a conditional flow-based pose prior for \emph{sparse inputs} which
builds upon prior work by: (1) generating high-quality 3D poses from an extremely sparse conditioning signal, 
(2) providing
\emph{latent variable sampling}, by learning a mapping from the observation to the region in the latent space that generates a likely and plausible pose.
This gives us a \emph{strong predictive model} as well an \emph{efficient pose prior} for optimization.
Furthermore, we
show that, in a conditional scenario, starting with $z=\mathbf{0}$ as in Kolotouros et al.~\cite{kolotouros2021probabilistic} does not necessarily lead to the best predictive outcome and that our method provides a more promising alternative.

\section{Preliminaries}
\label{sec:background}

\noindent\textbf{Normalizing Flows.}
\label{subsec:normalizingflow}
Normalizing flows~\cite{rezende2015variational} as likelihood-based generative models provide a path to an expressive probability distribution of the data.
Unlike VAEs, where the main challenge is to find an appropriate approximate posterior distribution, normalizing flows only require a definition of a simple base distribution (also referred to as a prior) and a series of bijective transformations.
These bijective transformations allow the model to map the data to the latent space and vice versa.

Given data $x\in\mathbb{R}^d$, the goal is to learn the joint distribution of data.
Normalizing flows model $x$ as a transformation~$T$ of a real vector $z\in\mathbb{R}^d$ sampled from the chosen base distribution $p_z(z)$, which could be as simple as a multivariate normal distribution.
With invertible and differentiable~$T$ (and hence $T^{-1}$) and using the change of variable formula~\cite{change_of_variable}, we obtain the density of $x$ as:
\begin{align}
\label{eq:change_of_variable1}
    p_x(x) = p_z(z) \big|\det J_T(z)\big|^{-1}
\end{align}
where $J_T$ is the Jacobian of $T$. Since $z=T^{-1}(x)$, $p_x(x)$ can also be written in terms of $x$ and the Jacobian of $T^{-1}$:
\begin{align}
\label{eq:change_of_variable2}
    p_x(x) = p_z(T^{-1}(x)) \big|\det J_{T^{-1}}(x)\big|
\end{align}
Instead of one transformation, multiple simple transforms can be composed to form a complex transform
$T=T_K\circ T_{K-1}\circ ... \circ T_1$ where $T_i$ transforms $z_{i-1}$ into $z_i$, $z_0$ is the latent variable in the base distribution and $x=z_K$. 
This composition can be built with neural networks 
that maximize the data log-likelihood. 
As shown in~\cite{rezende2015variational}, $\log p(x)$ can be written as:
\begin{align}
    \label{eq:flow}
    \log p(x) = \log p(z_0) - \sum_{i=1}^K \log \det \left|\frac{\partial T_{i}}{\partial z_{i}}\right|
\end{align}

\noindent\textbf{SMPL Body Model.}
\label{subsec:SMPL}
SMPL~\cite{loper2015smpl} is a parametric generative model of human body meshes.
SMPL receives as input the 3D human poses in axis-angle representation~$\theta$ and the body shape parameters~$\beta$, and generates the body mesh represented as $3\times6890$ matrix $M = \texttt{SMPL}(\theta, \beta)$. With that, we define $\texttt{SMPL}(\theta, \beta).\texttt{HH()}$ to compute the head and hands location and orientation.

\section{Proposed Method}
\label{sec:method}
We first define our problem statement, followed by an overview of our approach.
We then describe the components of FLAG and the training and generation of full body poses.

\subsection{Model Overview}
\label{subsec:problem_definition}
\label{subsec:overview}
Our task is to generate a full body pose $x_\theta$ given a sparse observation $x_{\mathbb{H}}$ and the shape parameters $\beta$.
$x_\theta\in\mathbb{R}^{3\times J}$ represents joint rotations as axis-angle vectors for $J$ body joints, and $x_{\mathbb{H}}\in\mathbb{R}^{9\times K}$ represents the global 6D joint rotation ~\cite{zhou2019continuity}
and a 3D joint location for each of the $K = 3$ observations (head and hands).
This information can be obtained from a parametric model of human body, e.g., SMPL~\cite{loper2015smpl}.

\begin{figure}
    \centering
    \includegraphics[width=\columnwidth]{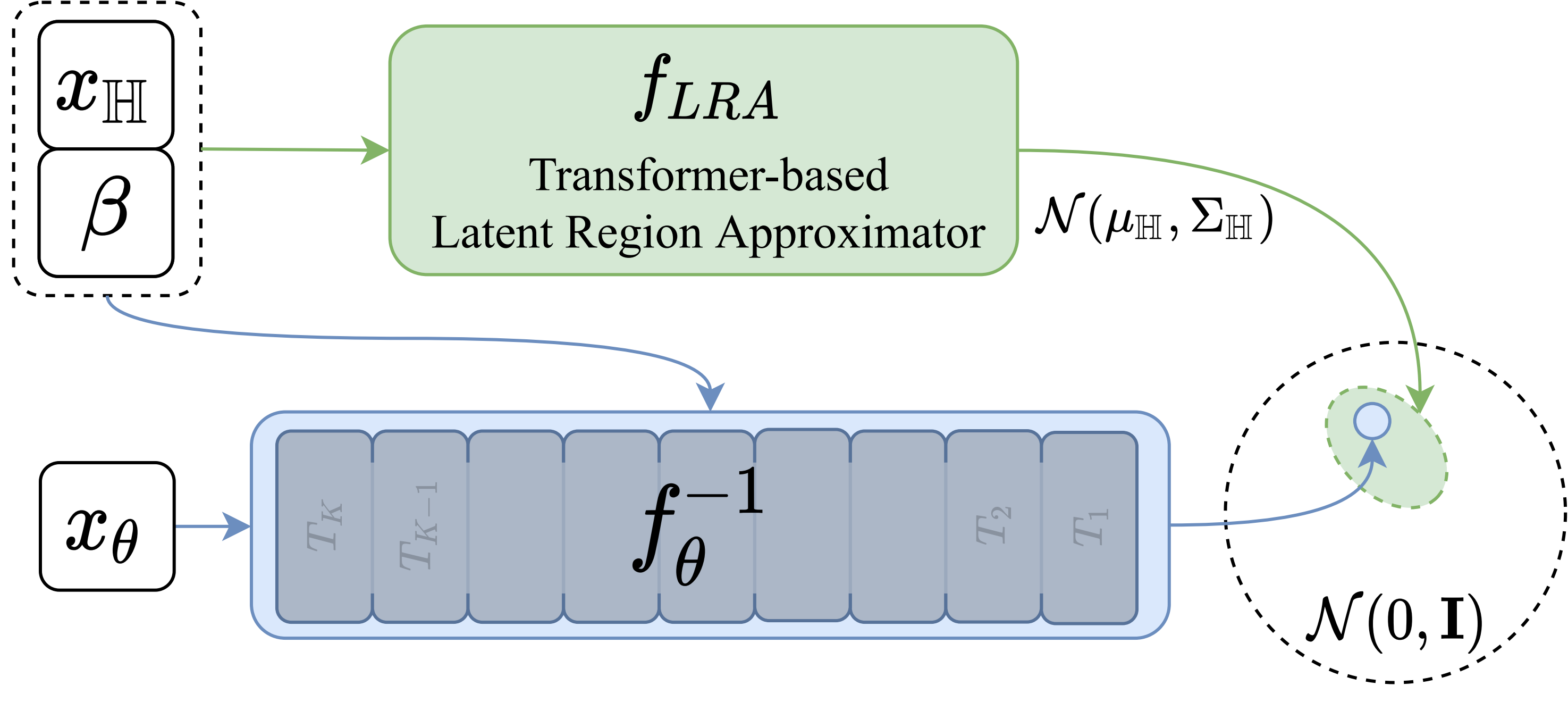}\vspace{-5pt}
    \caption{Overview of FLAG, consisting of a flow-based model $f_\theta$ and a latent region approximator $f_{\text{LRA}}$. During training $f_\theta$ aims to learn the distribution of $x_\theta$ and $f_{\text{LRA}}$ aims to learn a mapping from the condition to the latent representation of $x_\theta$.
    At test time, we sample a latent variable $z_\mathbb{H}$ via $f_{\text{LRA}}$ and use that to generate a new pose via $\hat{x}_\theta=f_\theta(z_\mathbb{H},[x_\mathbb{H},\beta])$.
    }
    \vspace{-10pt}
    \label{fig:method}
\end{figure}

One valid way to generate $x_\theta$ from $x_{\mathbb{H}}$ is to learn the distribution of the body pose given the observed $x_{\mathbb{H}}$ and $\beta$ via a conditional flow-based model $f_\theta$. 
While this approach can effectively provide the likelihood of a given pose, the generation process remains incomplete; for generating a novel pose given $x_{\mathbb{H}}$ and $\beta$, one needs to sample a latent variable.
However, the sampling process is completely independent of the observations. 
While~\cite{kolotouros2021probabilistic} rely on the mean of the latent space $z=\mathbf{0}$ (a vector of all zeros) as the latent code to generate the full pose, we argue that there exists a latent code that represents $x_\theta$ better than $z=\mathbf{0}$. In fact, while $z=\mathbf{0}$ is the most likely latent code in the base distribution, it may not necessarily translate to the most likely pose in the pose space since there can be changes in the volume of the distribution (the second term in Eq.~\ref{eq:flow}) through $f_\theta$'s transformations.
To obtain such a latent code, our model estimates a sub-region in the normalizing flow base distribution, $\mathcal{N}(\mu_\mathbb{H}, \Sigma_\mathbb{H})$, given $x_{\mathbb{H}}$ and $\beta$, from which a latent variable can be sampled to generate the full body pose.

At test time, to generate a full body pose given $x_{\mathbb{H}}$ and $\beta$, we sample a latent code from $z_\mathbb{H}\sim\mathcal{N}(\mu_\mathbb{H}, \Sigma_\mathbb{H})$ and use that as an approximation of $z_\theta$, the latent code that generates a full body pose.
We use this latent estimate to generate a full body pose via $\hat{x_\theta} = f_\theta(z_\mathbb{H},[x_{\mathbb{H}},\beta])$.
Next, we define $f_\theta$ and describe how we model $\mathcal{N}(\mu_\mathbb{H}, \Sigma_\mathbb{H})$.

\subsection{Flow over Full Body Pose}
\label{subsec:flow}
We model the distribution of $x_\theta$ with a normalizing flow model.
Our model $f_\theta$ is a conditional RealNVP~\cite{dinh2016density} conditioned on $x_{\mathbb{H}}$ and $\beta$.
This can be achieved by mapping $x_\theta$ from the pose distribution to the base distribution (and vice versa) via a composition of simple invertible transformations, where each transformation can stretch or shrink its input distribution.

\begin{figure}
\scriptsize
    \centering
    
    \begin{subfigure}[b]{\columnwidth}
         \centering
         \includegraphics[width=\columnwidth]{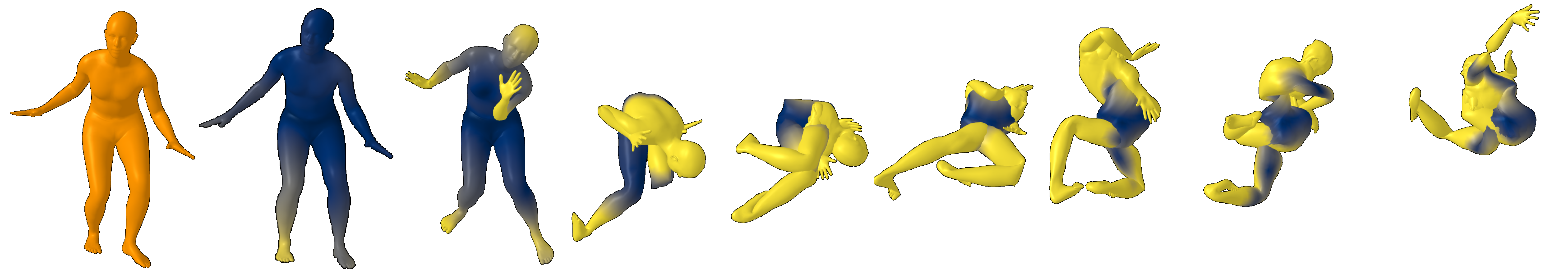}
         \caption{Without intermediate supervision}
         \label{fig:wo_intermediate}
     \end{subfigure}
    
    \begin{subfigure}[b]{\columnwidth}
         \centering
         \includegraphics[width=\columnwidth]{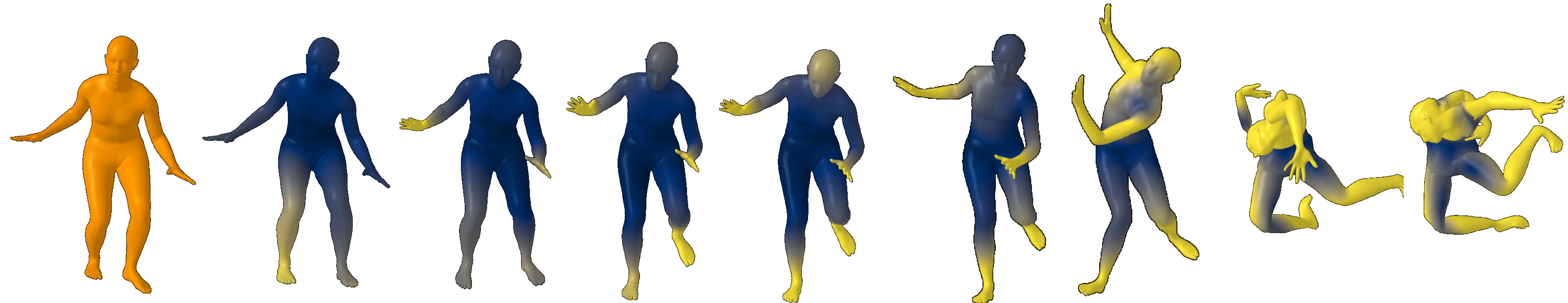}
         \caption{With intermediate supervision}
         \label{fig:w_intermediate}
     \end{subfigure}
         \vspace{-5pt}
    \caption{Pose progression through transformations from the base distribution to the pose distribution ($T_K \leftarrow T_1$) with and without intermediate supervision. First column shows the GT pose. The poses are color-coded to show large errors from GT in yellow, with dark blue showing zero error.}
    \vspace{-10pt}
    \label{fig:intermediate_supervision}
\end{figure}

While it is not straightforward to investigate the contribution of each invertible transformation in generating a human pose given $z_\theta$ sampled from the base distribution, we expect each successive transformation to add expressivity to the incoming distribution of human poses it acts upon.
To intuitively understand the role of each transformation, we visualize how a human pose is formed through all transformations in a model. 
Fig.~\ref{fig:wo_intermediate} 
illustrates how $z_\theta$ from the base distribution evolves through invertible transformations of $f_\theta$, up to the last transformation which produces a pose from the pose distribution.
As shown, most of the observable modifications to the intermediate distributions are happening in later stages, in which one observes a human-like pose being formed.
We argue that this is because the only source of supervision is the ground truth (GT) pose that explicitly guides the last transformation block.
To ease the training and get the most out of each transformation block in $f_\theta$, we propose to introduce \emph{intermediate supervision} to $f_\theta$.
In addition to having GT pose as the input to the last transformation block, we provide the GT pose as the input
to the intermediate transformation blocks, as if they are the last block of a sub-network.
This is possible because the transformations in $f_\theta$ do not modify the data dimension.
As a result, intermediate transformation blocks are encouraged to produce reasonable human poses and their capacity is exercised fully.
We illustrate the pose evolution through transformations with and without intermediate supervision in Fig.~\ref{fig:w_intermediate}) and also show in \ref{tab:intermediate_supervision} that intermediate supervision leads to improved plausibility of the generated poses.

\subsection{Latent Variable Sampling}
\label{subsec:transformer}
To generate a novel pose given $x_{\mathbb{H}}$ and $\beta$, we need to sample a latent variable $z$ from the base distribution and use that to generate a pose $\hat{x}_\theta=f_\theta(z,[x_{\mathbb{H}},\beta])$.
In a standard conditional flow-based model, one randomly samples $z~\sim\mathcal{N}(0, \mathbf{I})$, hoping for a plausible pose to be generated by the model, or consider $z=\mathbf{0}$~\cite{kolotouros2021probabilistic}.
Although these approaches yield valid solutions, we argue and empirically show that these do not constitute the best solution.
This can be examined explicitly thanks to the invertibility of normalizing flows, where one can obtain the oracle latent code $z^*=f_\theta^{-1}(x_\theta,[x_{\mathbb{H}},\beta])$.
Since an oracle latent code is known during training, we train our model such that it learns to map the condition ($x_{\mathbb{H}}$ and $\beta$) into a region in the base distribution where $z^*$ has a high likelihood. Utilizing $z^*$ during training allows us to take into account the changes in the volume of the base distribution, and thus the changes in the probability mass around the latent code, when transformed from the base distribution to the pose space via $f_\theta$.
We model the region of interest with a Gaussian and learn its parameters $\mu_\mathbb{H}$ and $\Sigma_\mathbb{H}=\text{diag}(\sigma_\mathbb{H})^2$. 
Such a mapping should have two desirable properties:
\textbf{(i)} It should be expressive, so that it can produce a representation of full body given the sparse observation. This is necessary to estimate a sub-region of the base distribution that represents the full body.
\textbf{(ii)} It should account for uncertainty in human body representation given sparse observation. When only the head and hands are observed, there exist multiple plausible full-body poses. For each plausible pose, we need to know the corresponding sub-region in the base distribution.
With these key properties in mind, we design a transformer-based mapping function with a discrete latent space.

\noindent\textbf{Attention-based Latent Region Estimation.}
We propose a transformer-based model (with a transformer encoder) to model the mapping function, taking advantage of the self-attention mechanism which learns the relationships between different joints in the body during training. 
Briefly, the transformer encoder receives as input $x_{\mathbb{H}}$ and $\beta$, and estimates $\mathcal{N}(\mu_\mathbb{H}, \Sigma_\mathbb{H})$ wherein $\mu_\mathbb{H}$ is trained to be a good approximation of the oracle latent code $z^*$. 

\begin{figure}
\scriptsize
    \centering
    \includegraphics[width=\columnwidth]{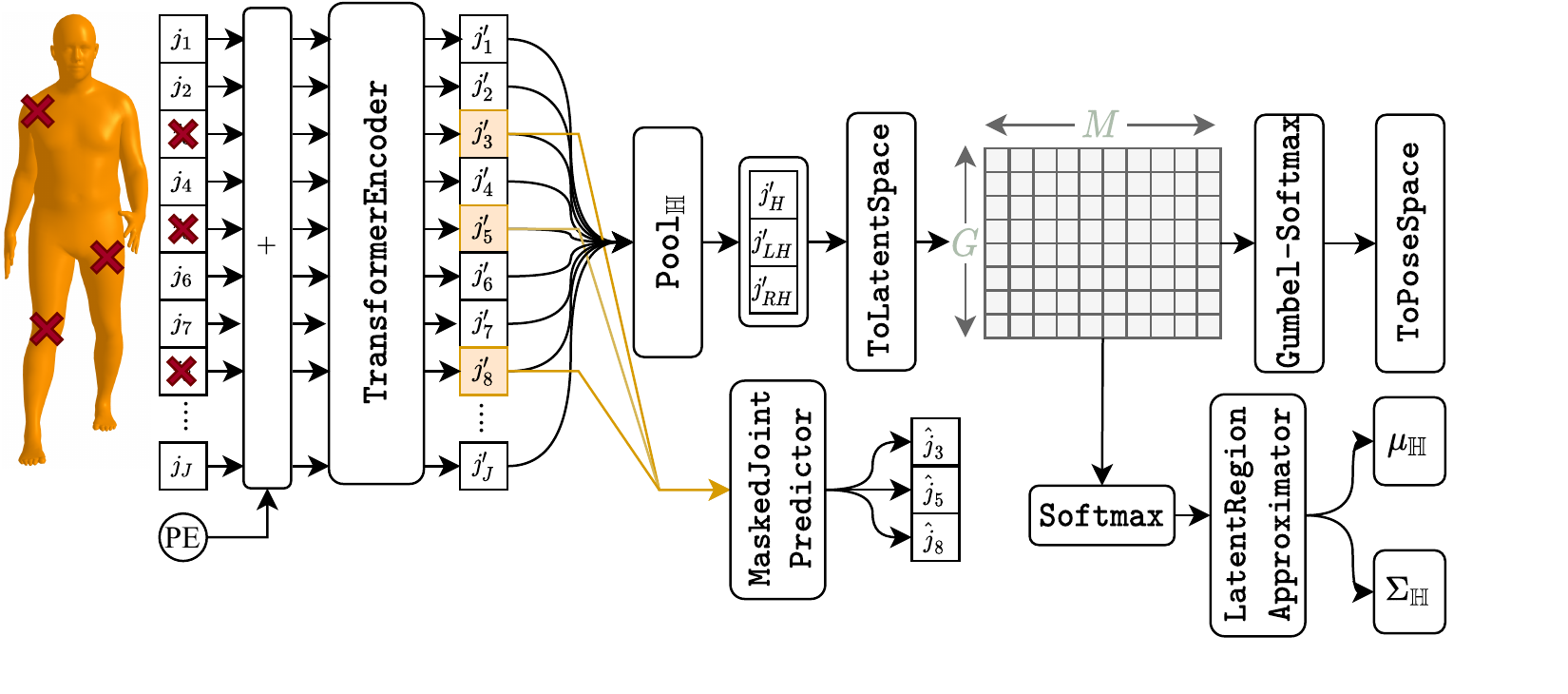}\vspace{-10pt}
    \caption{Transformer-based $f_{\text{LRA}}$. The attention-based encoder aims to learn the relationships between $x_\mathbb{H}$ and the rest of the body to generate an expressive representation of the body. A categorical latent space 
    from the output of the transformer encoder allows us to predict a plausible pose and the associated uncertainty.}
    \vspace{-10pt}
    \label{fig:transformer}
\end{figure}

For such a distribution to be representative of the full body, we make several design choices to come up with the model illustrated in Fig.~\ref{fig:transformer}.
First, training such a model using sparse inputs directly proves challenging.
To make it easier for the model, we define an auxiliary task of generating $x_\theta$ from the output of the transformer encoder (via \texttt{ToPoseSpace} block in Fig.~\ref{fig:transformer}), initially aiming to reconstruct $x_\theta$ from full body joints and gradually decreasing the joint visibility (through masking) in the encoding until we provide only head and hands\footnote{While in principle, masking can be done randomly, we follow the kinematic tree of the SMPL skeleton and start by masking the lower body joints, then the spine joints, followed by arm joints, and finally the pelvis (the root of the kinematic tree).}.
To further help the transformer learn the representation of the body, we introduce another auxiliary task of predicting the masked joints given the observed ones.
Such gradual masking-and-prediction (\texttt{MaskedJointPredictor}) lets the model infer the full body representation through attention (layer) on the available joints in the input.
To get a compact representation out of the transformer encoder, we apply  pooling ($\texttt{Pool}_\mathbb{H}$) over output joints and take only the head and hand representation, as they are always unmasked.

Next, we make the output of the transformer encoder stochastic, to obtain uncertainty estimates for the predicted poses.
We propose using a categorical latent space~\cite{jang2016categorical,richard2021meshtalk,saleh2021probabilistic} over human pose from the output of the transformer encoder (via \texttt{ToLatentSpace}).
One can sample a discrete latent variable from this distribution (via Gumbel-Softmax~\cite{jang2016categorical} for differentiablity) to generate $x_\theta$ with the defined auxiliary task or use the entire latent representation to estimate $\mathcal{N}(\mu_\mathbb{H}, \Sigma_\mathbb{H})$ (via \texttt{LatentRegionApproximator}) which contains information about a plausible pose and the associated uncertainty.
To efficiently model the complex distribution of human motion, we need a relatively large latent space, leading to a large number of latent categories.
To remedy this, we use a 2D categorical latent space, as shown in Fig.~\ref{fig:transformer}. We model a $G$-dimensional latent variable each responsible for $M$ modes, giving us a capacity to use $M^G$ one-hot latent codes.

\subsection{Learning}
We use a dataset 
of diverse 3D human models, where each sample is a triplet $(x_\theta, x_\mathbb{H}, \beta)$, where $\beta$ are the SMPL shape parameters. 
Our loss function $\mathcal{L}$ is given by
\begin{align}
\label{eq:loss}
    \mathcal{L} = & \lambda_{\text{nll}}\mathcal{L}_{\text{nll}} + 
                    \lambda_{\text{mjp}}\mathcal{L}_{\text{mjp}} +
                    \lambda_{\text{rec}}\mathcal{L}_{\text{rec}} + 
                    \lambda_{\text{lra}}\mathcal{L}_{\text{lra}}
\end{align}
where $\lambda_.$s are the weights associated with each term. 

\noindent\textbf{$\mathcal{L}_{\text{nll}}$:}
This term encourages the model to minimize the negative log-likelihood of $x_\theta$ under the model $f_\theta$, following Eq.~\ref{eq:flow}.
Additionally, we take into account the log-likelihoods produced by the sub-networks of $f_\theta$ as the result of intermediate supervision discussed in Section.~\ref{subsec:flow}. 
\begin{align}
    \mathcal{L}_{\text{nll}} = - \big(\log p_\theta(x_\theta) + 
                                \sum_{s\in S} w_s \log p_\theta^s(x_\theta)\big)
\end{align}
where $S$ is the set of sub-networks of $f_\theta$ (e.g., from block $T_i$ to $T_1$ for a pre-defined set of $i$s), $p_\theta^s(x_\theta)$ is the likelihood of $x_\theta$ under sub-network $s$, and $w_s$ is the weight associated to the sub-network that is proportional to the number of transformation blocks in each sub-network.

\noindent\textbf{$\mathcal{L}_{\text{mjp}}$:}
To train the auxiliary task of masked joint prediction, we employ the term
\begin{align}
    \mathcal{L}_{\text{mjp}} = \sum_{j\in J_{\text{masked}}}\left\|\hat{x}_P^j - x_P^j\right\|_2^2
\end{align}
where $J_{\text{masked}}$ is the list of masked joints,
$x_P^j$ is the representation of the $j^{th}$ joint in $\mathbb{R}^9$ (6D rotation and 3D location), and $\hat{x}_P^j$ is the corresponding prediction from the network.

\noindent\textbf{$\mathcal{L}_{\text{rec}}$:}
This term acts on the output of the auxiliary task of decoding the full body pose from a discrete latent variable sampled transformer's categorical latent space, aiming to guide to build a meaningful discrete latent space.
\begin{align}
    \mathcal{L}_{\text{rec}} = \left\|\hat{x}_\theta^{\text{tps}} - x_\theta\right\|_2^2
\end{align}
where $\hat{x}_\theta^{\text{tps}}$ is the output of \texttt{ToPoseSpace}.

\noindent\textbf{$\mathcal{L}_{\text{lra}}$:}
Finally, this term encourages learning a Gaussian distribution $\mathcal{N}(\mu_\mathbb{H},\Sigma_\mathbb{H})$ under which the oracle latent variable $z^*$ has high likelihood. 
\begin{align}
    \mathcal{L}_{\text{lra}} = -\alpha_{\text{nll}}\log p_\mathbb{H}(z^*) + \alpha_{\text{rec}} \left\|\mu_\mathbb{H} - z^*\right\|_2^2 \nonumber \\ - \alpha_{\text{reg}} (1 + \ln \sigma_\mathbb{H} - \sigma_\mathbb{H})  
    \label{eq:lra}
\end{align}
where $p_\mathbb{H}$ is the estimated sub-region of the base distribution.
While the first term in Eq.~\ref{eq:lra} is enough to achieve this goal, we add the second term to implicitly encourage $\mu_\mathbb{H}$ to be similar to $z^*$ and the third term discourages $\sigma_\mathbb{H}$ to be zero and thus avoids a deterministic mapping. Note that $\alpha_{\text{reg}}$ and $\alpha_{\text{rec}}$ can be relative small, but need to be present.

Although the entire model can be trained in an end-to-end fashion, we observed training $f_\theta$ first followed by training the latent region approximator is quite effective since we have access to a valid $z^*$ from the beginning.
The second training stage is quick, $~4$ GPU-hours.
This two-stage training may also be useful in cases where one wishes to use a previously trained $f_\theta$ as a foundation model~\cite{bommasani2021opportunities} and only train mapping functions for other data modalities, e.g., body markers or environment scans.

\begin{figure*}[!t]
    \centering
    \includegraphics[width=0.9\textwidth]{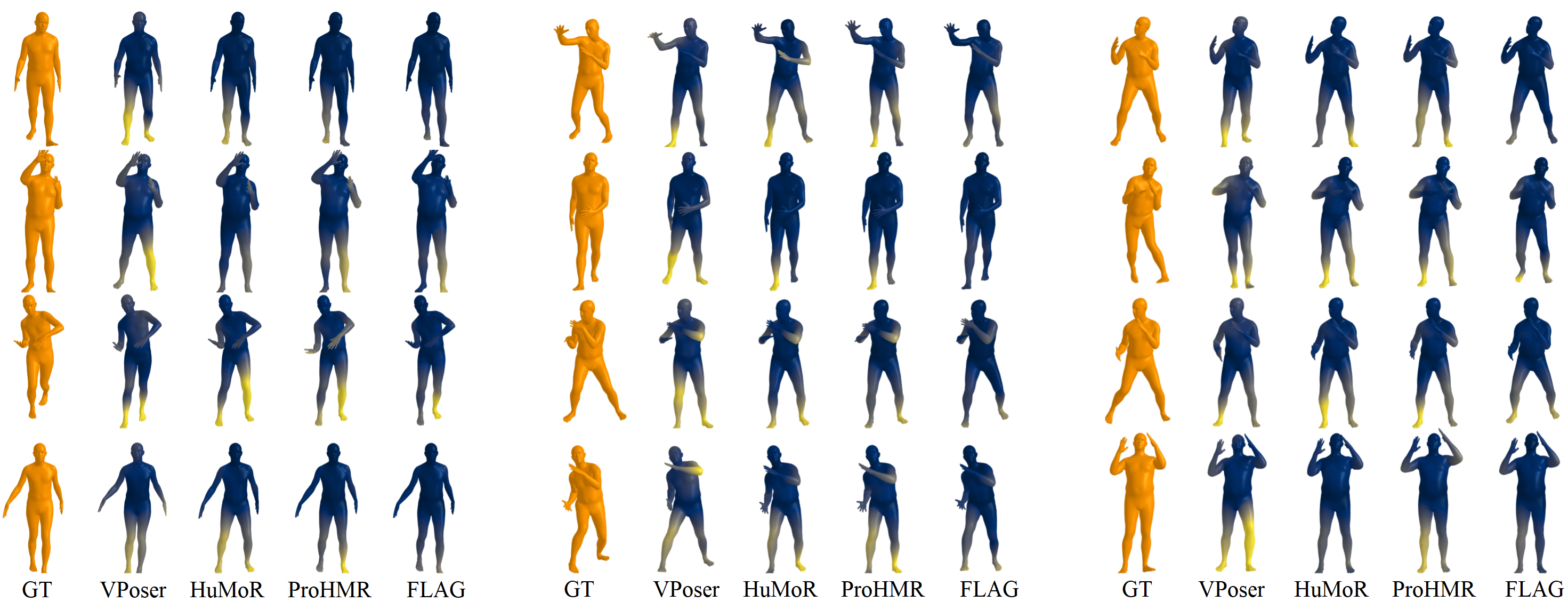}\vspace{-5pt}
    \caption{Qualitative results. First column (orange) shows the GT. Generated poses are color-coded to show large vertex errors in yellow.
    }
    \label{fig:qualitative}
    \vspace{-10pt}
\end{figure*}

\subsection{Conditional Generation}
\label{subsec:optimization}
We can generate a full body pose given $x_\mathbb{H}$ and $\beta$ by first computing $\mu_\mathbb{H}$ given the observation, then use $\mu_\mathbb{H}$ as an approximation of $z_\theta$ to generate a pose $\hat{x_\theta}=f_\theta(\mu_\mathbb{H}, [x_\mathbb{H}, \beta])$. To further enhance our quality of the generated pose, one can also use our flow-based model as a pose prior in optimization to minimize a cost function over the prior and the data. The optimization can be done either in pose space or in latent space. We use the LBFGS optimizer~\cite{liu1989limited} throughout (see supp.\ mat. for further details).

\noindent\textbf{Optimization in the pose space:} The optimizer seeks a plausible human pose $\theta$ under our model that matches the observation $x_\mathbb{H}$. We optimize for $\theta$ by minimizing the cost:
\begin{align}
    \mathcal{C}(\theta) = -\log p_{\theta}(x_\theta) + ||\texttt{SMPL}(\theta, \beta). \texttt{HH()} - x_\mathbb{H}||^2
    \label{eq:opt_theta}
\end{align}

\noindent\textbf{Optimization in the latent space:} The optimizer seeks a latent variable $z$ that leads to a plausible pose under the model that matches the observation $x_\mathbb{H}$. Using generative functionality of the pose prior ($f_\theta$) to generate a pose, we optimize for $z$ by minimizing the cost:
\begin{align}
    \mathcal{C}(z) = -\log p(z) + ||\texttt{SMPL}(\hat{\theta}, \beta).\texttt{HH()} - x_\mathbb{H}||^2 + r
    \label{eq:opt_z}
\end{align}
where $\log p(z)$ is the log-likelihood of the optimized $z$ under the base distribution $\mathcal{N}(0, \text{I})$, $\hat{\theta}=f_\theta(z, [x_\mathbb{H}, \beta])$, and $r=||z - \mu_{\mathbb{H}}||$ is a regularizer (see supplementary material) to implicitly prevent the latent code from straying too far from the initial guess (there is no signal for the lower body in the data term, i.e., the second term in Eq.~\ref{eq:opt_z}).

\section{Experiments}
\label{sec:experiment}

We first introduce the dataset and then present the experimental results, ablation studies, and qualitative results of our approach (see supp.\ mat. for implementation details).

\noindent\textbf{Dataset.}
We report results on AMASS~\cite{AMASS:ICCV:2019}, a large-scale motion capture dataset, with diverse poses represented with SMPL body model. We evaluate our approach and existing methods on the held out Transitions and HumanEVA~\cite{sigal2010humaneva} subset of the AMASS. The models are trained on the remaining datasets, excluding the dancing sequences~\cite{rempe2021humor}.

\begin{table}
\setlength{\tabcolsep}{10pt}
\scriptsize
    \centering
    \begin{tabular}{l c c}
    \toprule
    Method & Upper Body MPJPE ($\downarrow$) & Full Body MPJPE ($\downarrow$) \\
    \midrule
    VPoser-HMD & 1.69 cm & 6.74 cm \\
    HuMoR-HMD & 1.52 cm & 5.50 cm\\
    VAE-HMD & 3.75 cm & 7.45 cm\\
    ProHMR-HMD & 1.64 cm & 5.22 cm\\
    \midrule
    FLAG (Ours) & \textbf{1.29 cm} & \textbf{4.96 cm} \\
    \bottomrule
    \end{tabular}
    \vspace{-5pt}
\caption{Comparison of FLAG with existing methods on AMASS.}\vspace{-10pt}
\label{tab:comparison}
\end{table}

\noindent\textbf{Baselines.}
There have been a few efforts towards generating full body poses given head and hand inputs~\cite{dittadi2021full}. Our first baseline, which we call \textbf{VAE-HMD}, involves a two-step process. First a VAE encoder-decoder is trained on full body, without any condition. In the next step, another VAE is trained (starting from the frozen decoder) which encodes head and hand representation into the latent space and uses the previously trained full body decoder for generation. Since our approach is a conditional pose prior, we compare it with existing conditional pose priors after adapting them to our problem setting. ProHMR~\cite{kolotouros2021probabilistic} is closest to our approach in terms of architecture since it is a conditional flow-based model. We adapt the conditioning signal to head and hands representation and include this as another baseline called \textbf{ProHMR-HMD}. Our third baseline is a conditional version of VPoser~\cite{pavlakos2019expressive}, a VAE-based approach, since it constitutes a strong and commonly used human pose prior in the literature, and refer to it as \textbf{VPoser-HMD}. We also evaluate another recently proposed CVAE-based pose prior, HuMoR~\cite{rempe2021humor}, which learns a prior distribution given the conditioning signal. We adapt this approach to our scenario and refer to it as \textbf{HuMoR-HMD}. 
For all baselines we follow original implementation where available, otherwise follow the papers. We consider the same data and condition representations for all methods for a fair comparison.
Following prevailing convention \cite{rempe2020contact,dittadi2021full}, the avatar root is positioned at the origin. 

\noindent\textbf{Evaluation Metrics.}
To measure accuracy quantitatively, we report the mean per-joint position error (MPJPE) in cm. Since the quality of upper-body representation is of greater importance for AR, VR, and MR applications, we report the MPJPE of the upper body as well as that of the full body.

\subsection{Comparison to Existing Approaches}
We evaluate our approach in generating a plausible pose given sparse observations and compare it with existing methods. Table~\ref{tab:comparison} summarizes this evaluation\footnote{The MPJPE for VAE-HMD on the standard AMASS test set is relatively high. We analyze this in the supplementary material, demonstrating that this is due to imperfect utilization of the latent space resulting from the two-stage training used in VAE-HMD approach.}. We do not use optimization for this comparison. Flow-based approaches, ProHMR-HMD and FLAG (Ours), generally yield lower full body error, but  approaches that have conditional latent variable sampling tend to generate better upper bodies. This is the case with HuMoR and our approach, where the latent variable is sampled given head and hands information while for other techniques, a latent variable is sampled independent of the conditioning signal. The superiority of our approach is also evident in the qualitative results in Fig.~\ref{fig:qualitative}, where FLAG yields least error compared to other techniques, with HuMoR producing relatively good upper body. We provide more qualitative results in the supplementary material.

\subsection{Ablation Study}

\noindent\textbf{Effect of Intermediate Supervision.}
Building on Section~\ref{subsec:flow}, here we evaluate the effect of the proposed intermediate supervision. Such supervision provides an additional signal to intermediate transformation blocks of $f_\theta$, allowing better convergence to a plausible pose throughout transformations when starting from the base distribution. This was visible in Fig.~\ref{fig:intermediate_supervision} and is also evident in our quantitative results in Table~\ref{tab:intermediate_supervision} wherein we show considerable improvement in the quality of generated poses.

\begin{table}
\setlength{\tabcolsep}{5pt}
\scriptsize
    \centering
    \begin{tabular}{l c c}
    \toprule
    Setting & Upper Body MPJPE ($\downarrow$) & Full Body MPJPE  ($\downarrow$) \\
    \midrule
    w/o Intermediate Sup. & 1.64 cm & 5.22 cm \\
    w/ Intermediate Sup. & \textbf{1.39 cm} & \textbf{5.11 cm} \\
    \bottomrule
    \end{tabular}
\vspace{-5pt}
    \caption{Evaluating the effect of intermediate supervision.}\vspace{-10pt}
    \label{tab:intermediate_supervision}
\end{table}

\begin{table}
\scriptsize
    \centering
    \begin{tabular}{l c c c}
    \toprule
     Method & GT Pose & Manip. Pose (RD $\uparrow$) & Noise (RD $\uparrow$)\\
     \midrule
     CVAE * (true ELBO)  & 29.68 & 29.68 (0.0) & 32.40 (0.08) \\
     VPoser-HMD * & 34.79 & 35.56 ( 0.02) & 2.39$\times 1e3$ (0.98)\\
     HuMoR-HMD * & 46.02 & 49.21 (0.06) & 2.37$\times 1e4$ (0.99)\\
     ProHMR-HMD \dag & 110.72 & 282.01 (0.61) & 6.63$\times 1e7$ \textbf{(1.0)}\\
     \midrule
     FLAG (Ours) \dag & 98.54 & 489.66 \textbf{(0.80)} & 3.04$\times 1e13$ \textbf{(1.0)}\\
    \bottomrule
    \end{tabular}
\vspace{-5pt}
\caption{Evaluating the generalizability of learned latent representation by examining the NLL of in- and out-of-distribution samples. VAE-based methods are denoted with * while the NF-based ones are denoted with \dag.
    }\vspace{-10pt}
    \label{tab:ood}
\end{table}

\noindent\textbf{Generalizability of Latent Representations.}
As various models are trained differently, with various training tricks such as KL term annealing or modifying the ELBO for mitigating posterior collapse, we define an auxiliary task to evaluate the quality of the learned latent space. To this end, we use the negative log-likelihood (NLL) metric to identify out-of-distribution (OOD) samples. We define in-distribution samples to be the poses from the ground truth test set, whereas the OOD samples are defined in two ways (1) manipulating ground truth poses by adding a small amount of noise to a subset of joints (2) creating pose-like random noise (random values within the range of natural poses). Table~\ref{tab:ood} summarizes how different models perform when detecting OOD samples. For a clearer comparison, we also report the relative difference ($\text{RD}=\frac{|\text{NLL}_{\text{OOD}}-\text{NLL}_{\text{GT}}|}{\max(\text{NLL}_{\text{OOD}}, \text{NLL}_{\text{GT}})}$) between the NLL of the models for OOD samples and that of the ground truth poses, which higher is better. It can be seen that flow-based models are typically better at detecting OOD samples, demonstrating a richer learned latent representation, whereas VAE-based ones are less effective despite utilizing various techniques to avoid posterior collapse. For reference, we also provide the results of a CVAE trained with true ELBO.

\begin{table}
\setlength{\tabcolsep}{5pt}
\scriptsize
    \centering
    \begin{tabular}{l c c}
    \toprule
    Latent Variable Sampling & Upper Body MPJPE ($\downarrow$) & Full Body MPJPE  ($\downarrow$) \\
    \midrule
    Zeros ($z=\mathbf{0}$) & 1.39 cm & 5.11 cm \\
    MLP ($z=\text{MLP}_\mathbb{H}$) & 1.36 cm& 5.05 cm \\
    Ours ($z=\mu_\mathbb{H}$) & \textbf{1.29 cm} & \textbf{4.96 cm} \\
    \bottomrule
    \end{tabular}
\vspace{-5pt}
\caption{Evaluating the effect of latent variable sampling. Comparing $z=\mathbf{0}$~\cite{kolotouros2021probabilistic}, estimating $z$ with an MLP, and our approach.}\vspace{-10pt}
    \label{tab:sampling}
\end{table}

\begin{table}
\setlength{\tabcolsep}{12pt}
\scriptsize
    \centering
    \begin{tabular}{l c c}
    \toprule
    Latent Variable Sampling & Cosine Dist.($\downarrow$) & Sinkhorn Dist.($\downarrow$) \\
    \midrule
    Random ($z\sim\mathcal{N}(0, \mathbf{I})$) & 1.0 & 0.29\\
    Zeros ($z=\mathbf{0}$) & 1.0 & 0.22 \\
    Ours ($z=\mu_\mathbb{H}$) & \textbf{0.81} & \textbf{0.18} \\
    \bottomrule
    \end{tabular}
\vspace{-5pt}
\caption{Distance to oracle latent code $z^*=f_\theta^{-1}(x_\theta)$.}\vspace{-10pt}
\label{tab:oracle_z_distance}
\end{table}

\begin{figure}
    \centering
    \includegraphics[width=0.35\textwidth]{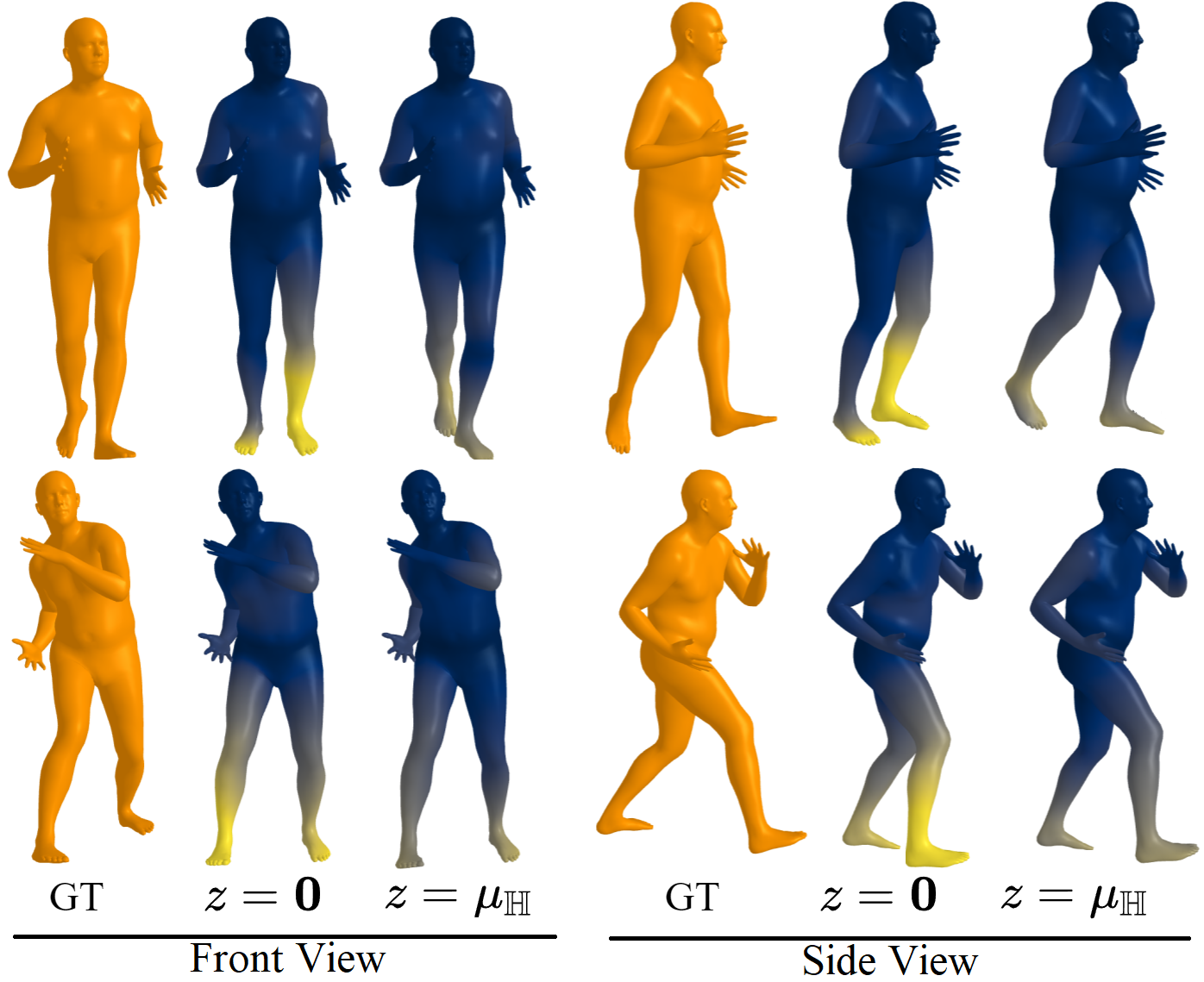}\vspace{-5pt}
    \caption{Qualitative evaluation of latent variable sampling, comparing our prediction from $z=\textbf{0}$ and from $z=\mu_\mathbb{H}$. Generated poses are color-coded to show large vertex errors in yellow.}
    \label{fig:init_z}
    \vspace{-10pt}
\end{figure}

\noindent\textbf{Effect of Initial Latent Code.}
A key contribution of this work is the probabilistic mapping from the condition to a sub-region in the latent space that leads to a highly plausible pose. In Table~\ref{tab:sampling}, we compare our approach, $z=\mu_\mathbb{H}$ with the proposal of ProHMR-HMD~\cite{kolotouros2021probabilistic} which claims $z=\mathbf{0}$ yields the most plausible pose. While $z=\mathbf{0}$ yields \emph{a} plausible pose, this experiment shows the existence of a better latent code, $z_\mathbb{H}$ that leads to a more plausible pose that has a high likelihood under the model. This is also shown in Table~\ref{tab:oracle_z_distance}, where we compute the distance between the oracle latent code $z^*=f_\theta^{-1}(x_\theta, [x_\mathbb{H}, \beta])$ to the latent code from our approach as well as that of ~\cite{kolotouros2021probabilistic}. For the sake of completeness, we also compare our method with an MLP that learns to find a good latent code given the condition, which we refer to as $z=\text{MLP}_\mathbb{H}$ in Table~\ref{tab:sampling}. In addition to quantitative evaluation, Fig.~\ref{fig:init_z} shows the effect of a proper initial latent code in generating pose from sparse observation.

\begin{figure}
    \centering
    \includegraphics[width=\columnwidth]{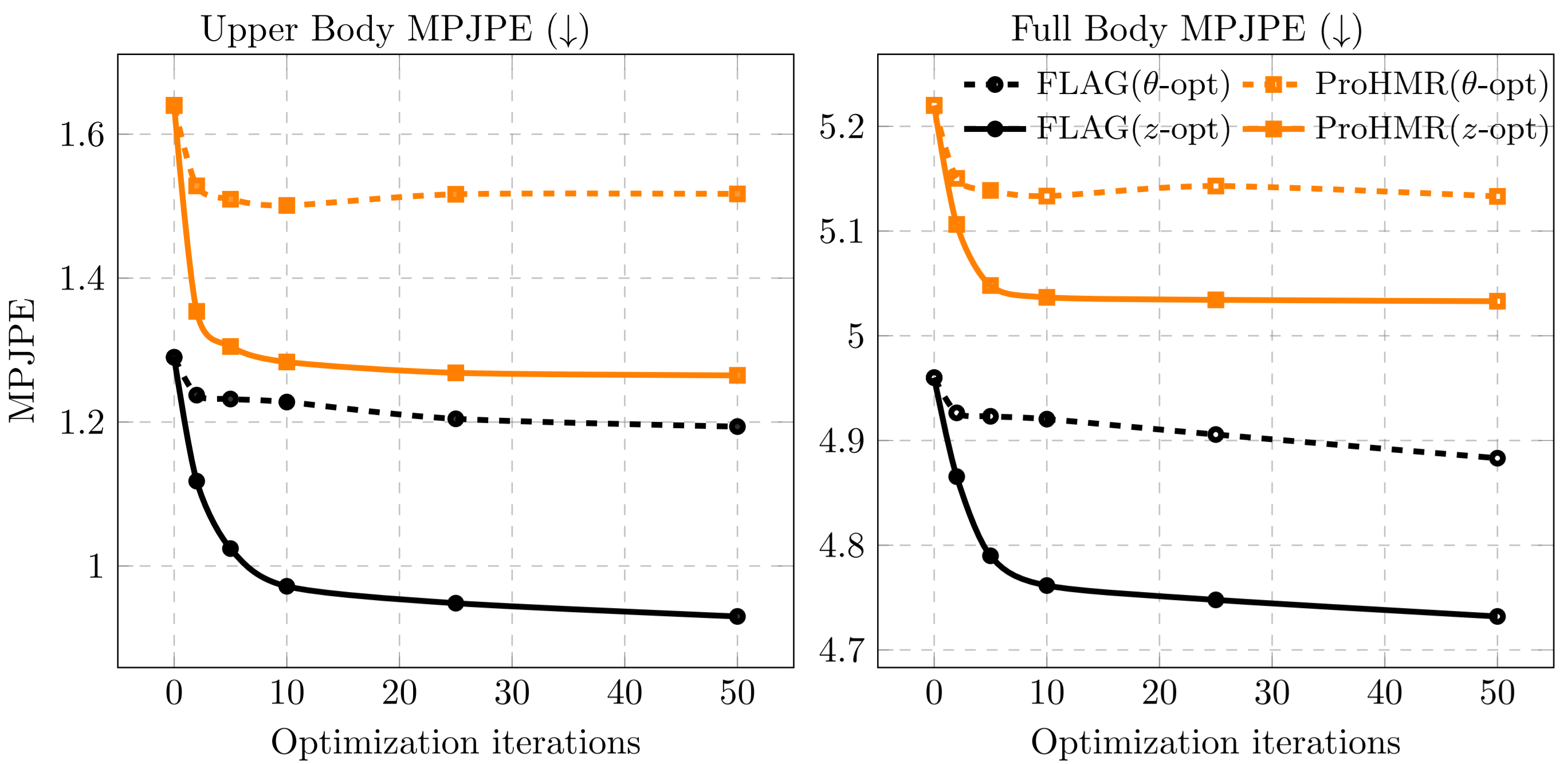}\vspace{-5pt}
    \caption{MPJPE as a function of optimization iteration. As shown here, ProHMR-HMD requires 50 iterations of optimization in the latent space to achieve an error on par with our approach with a proper initial latent code $z=\mu_\mathbb{H}$ without any optimization. Optimizing in the latent space yields lower error compared to optimizing in the pose space.}
    \label{fig:optimization_result}
    \vspace{-15pt}
\end{figure}

We also observed that initial latent variable affects the quality of predicted poses refined via optimization in either the pose space or the latent space, as described in Section~\ref{subsec:optimization}. We evaluate this in Fig.~\ref{fig:optimization_result}, where we use flow-based approaches as a pose prior in the optimization process and report the MPJPE. Consistent with Table~\ref{tab:sampling}, the results demonstrate that a proper initialization leads to a better performance. Given a fixed optimization budget, our method reaches a desired error threshold quicker owing to (a) a better initialization and (b) more reliable likelihood estimates (supported by results in~\ref{tab:ood}). For instance, even after 50 iterations of optimization, ProHMR-HMD~\cite{kolotouros2021probabilistic} does not outperform the solution reached by our approach after 2 optimization iterations regardless of the (pose or latent) space we optimize in. Finally, we also demonstrate that optimization in the latent space generally yields lower error compared to optimization in the pose space, for either model designs. 

\begin{figure}
    \centering
    \includegraphics[width=0.85\columnwidth]{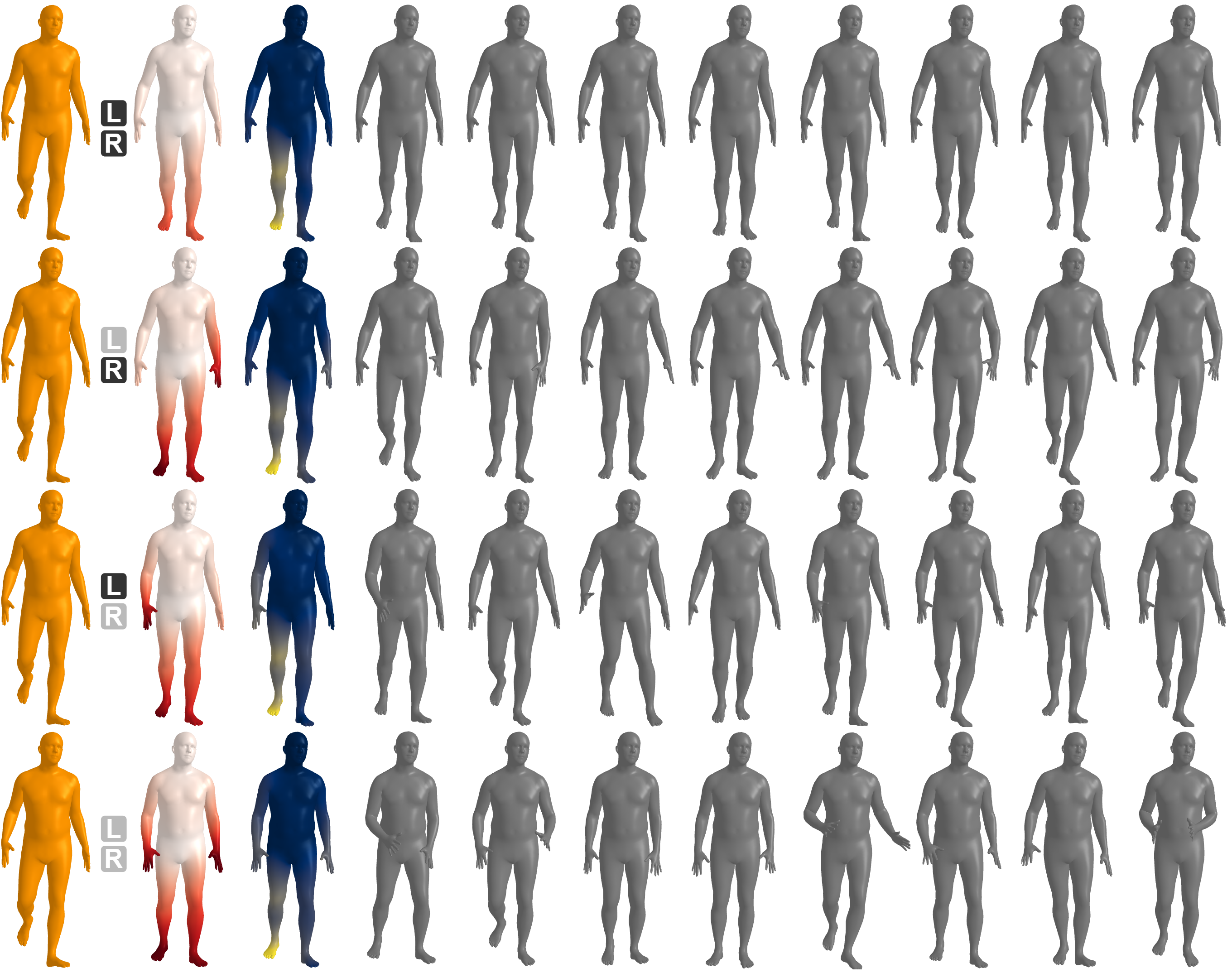}\vspace{-5pt}
    \caption{Qualitative results of FLAG when dealing with partially visible hands. From left to right, we illustrate the GT, the avatar's hand visibility status (black box is visible, gray box is invisible), uncertainty map on the pose from $z_\mathbb{H}=\mu_\mathbb{H}$ colorized based on the uncertainty (white is certain, red is uncertain), the pose from $\mu_\mathbb{H}$, followed by generated poses starting with samples from $\mathcal{N}(\mu_\mathbb{H},\Sigma_\mathbb{H})$. 
    }
    \label{fig:partial_visibility}
    \vspace{-10pt}
\end{figure}

\noindent\textbf{Partial Hand Visibility.}
Our method, as well as all other baselines, work on the assumption that head and hand signals are always visible. While this is the case with the head signal in practice, both hands may not always be visible, if they go out of the field of view. In real-world applications, robustness to hands going out of field of view is important, so we evaluate the performance of our method under partial or no hand visibility. The use of aggressive joint masking in our model allows us to fine-tune our model for another 10 epochs and randomly mask hands with probability of $p=0.2$. In Fig.~\ref{fig:partial_visibility}, we demonstrate that FLAG can generate highly plausible poses under partial or no hand observations\footnote{To get the uncertainty maps in Fig.~\ref{fig:partial_visibility}, we generate $K$ poses from $z\sim\mathcal{N}(\mu_\mathbb{H},\Sigma_\mathbb{H})$ and compute the vertices' distance of these sampled poses from the one generated with $z=\mu_\mathbb{H}$.}. 

\noindent\textbf{Limitations and Future work.}
While FLAG is capable of generating highly plausible poses given extremely sparse observations in the majority of scenarios, it may fail to generate complex, less common lower-body poses, e.g., martial arts (examples are provided in the supp.\ mat.), potentially because these poses are not very common in the training dataset.
FLAG uses only static pose information; extending FLAG to consume temporal data is a natural research direction.
We use only HMD signal as the input to the model, whereas in some AR/VR scenarios, other modalities such as audio or environment scans may also be available.
Although FLAG aims to find a better latent code to generate a plausible pose, there may still be a considerable gap between our estimated latent code and the oracle one (see Table~\ref{tab:oracle_z_distance}).
Further exploration in this area may lead to more faithful and accurate avatar poses.

\noindent\textbf{Societal impact.}
While current datasets such as AMASS have a large number of poses, the data comes from 346 subjects who may not represent the true diversity of the global population. 
We have more work to do as a community to represent people of all age groups, and people with disabilities (\eg wheelchair users, amputees).
For anyone with a body morphology outside of the distribution represented in the datasets, we should ask: 1) Does the technology work for them? 2) Can they choose how they want to be represented?
There could be negative outcomes from representing an individual in a way that removes a disability from view, for example. Mixed reality applications bring the promise of enhanced remote collaboration and communication, but there may also be potential negative societal impacts: misrepresentation including impersonation,
further marginalization of socio-economically disadvantaged groups caused  unknowingly or intentionally.  
Even so, with mindful deployment of technology and appropriate governance, we remain positive that realistic human representations can help the world grow closer without the harmful environmental impact of long-distance travel.

\section{Conclusion}
\label{sec:discussion}
We presented FLAG, a new approach to generate plausible full body human poses from sparse HMD signals. 
FLAG is a conditional flow-based generative model of the 3D human body from sparse observations; we not only learn a conditional distribution of 3D human body, but also a probabilistic mapping from the observation to the latent space from which we generate plausible poses with uncertainty estimates.
We show that our approach is both a strong predictive model, and an efficient pose prior in different optimization settings, thanks to our latent variable sampling mechanism.
Experimental evaluation and ablation studies demonstrated that our method outperforms state of the art methods on the challenging AMASS dataset, requires fewer optimization iterations and leads to a very low error.

\noindent\textbf{Acknowledgements.}
We thank Tom Minka and Darren Cosker for comments that greatly improved the manuscript.

\newpage
{\small
\bibliographystyle{ieee_fullname}
\bibliography{flag}
}

\end{document}